\documentclass{tlp}
\usepackage{aopmath}

\def\beq{\begin{equation}}
\def\eeq#1{\label{#1}\end{equation}}
\def\ba{\begin{array}}
\def\ea{\end{array}}
\def\o{\overline}

\def\i#1{\hbox{\it #1\/}}
\def\ar{\leftarrow}
\def\Ar{\Leftarrow}
\def\rar{\rightarrow}
\def\lrar{\leftrightarrow}
\def\llrar{\Leftrightarrow}

\def\no{\i{not}}

\def\sm{\hbox{\rm SM}}
\def\tr{\hbox{\rm Tr}}

\def\bx{\hbox{\bf{x}}}

\def\bP{\hbox{\bf{p}}}
\def\bp{\hbox{\bf{p}}}
\def\v{\hbox{$\upsilon$}}
\def\bph{\widehat{\hbox{\bf{p}}}}

\def\pa{\hbox{\it Pos}}
\def\na{\hbox{\it Neg}}
\def\uc{\widetilde\forall}
\def\ouc{\widetilde\forall_{obj}}
\def\wh{\widehat}

\usepackage{amssymb}

\begin{document}

\title{Representing First-Order Causal Theories by Logic Programs}

\author[P. Ferraris, J. Lee, Y. Lierler, V. Lifschitz, and F. Yang]
{Paolo Ferraris\\
Google, Inc.\\
\email{otto@cs.utexas.edu}\\
\and
Joohyung Lee\\
School of Computing, Informatics and Decision Systems Engineering,\\
Arizona State University\\
\email{joolee@asu.edu}\\
\and
Yuliya Lierler\\
Computer Science Department, University of Kentucky\\
\email{yuliya@cs.utexas.edu}\\
\and
Vladimir Lifschitz and Fangkai Yang\\
Department of Computer Science, University of Texas at Austin\\
\email{\{vl,fkyang\}@cs.utexas.edu}\\
}

\pagerange{\pageref{firstpage}--\pageref{lastpage}}
%\volume{\textbf{10} (3):}
%\jdate{Feb 2011}
\setcounter{page}{1}
%\pubyear{2011}
\submitted{Aug, 16, 2010}
\revised{Feb, 21, 2011}
\accepted{Mar, 16, 2011}
\maketitle

\label{firstpage}

\begin{abstract}
Nonmonotonic causal logic, introduced by Norman McCain and Hudson Turner,
became a basis for the semantics
of several expressive action languages.  McCain's embedding of definite
propositional causal theories into logic programming paved the way to
the use of answer set solvers for answering queries about actions
described in such languages.  In this paper we extend this embedding to
nondefinite theories and to first-order causal logic.

\end{abstract}

\section{Introduction}

Propositional nonmonotonic causal logic \cite{mcc97} and its generalizations
became a basis for the semantics of several expressive action languages
\cite{giu98,giu04,lif06,lif07,ren09}. The Causal Calculator
({\sc CCalc})\footnote{\tt http://www.cs.utexas.edu/users/tag/ccalc/}
is a partial implementation of this logic that allows us to automate some
kinds of reasoning and planning in action domains described in such
languages.  It has been used to solve several challenging
commonsense reasoning problems, including problems of nontrivial size
\cite{akm04}, to provide a group of robots with high-level reasoning
\cite{erd09}, to give executable specifications of norm-governed
computational societies \cite{art09}, and to automate the analysis of
business processes under authorization constraints \cite{arm09}.

An important theorem due to Norman McCain \cite[Proposition~6.7]{mcc97c} shows
how to embed a fragment of propositional causal logic into the language of
logic programming under the answer set semantics \cite{gel91b}.  This result,
reviewed below, paved the way to the development of an attractive alternative
to {\sc CCalc}---the software system {\sc coala} \cite{geb10} that uses answer
set programming \cite{mar99,nie99,lif08} for answering queries about
actions described in causal logic.

A causal theory in the sense of \cite{mcc97} is a set of ``causal rules''
of the form $F\Ar G$, where~$F$ and~$G$ are propositional
formulas (the {\sl head} and the {\sl body} of the rule).  The rule reads
``$F$ is caused if~$G$ is true.''
Distinguishing between being true and having a cause turned out to be
essential for the study of commonsense reasoning.  The assertion
``if the light is on at time~0 and you toggle the switch then the light will be
off at time~1'' can be written as an implication:
$$\i{on}_0\land\i{toggle}\rar\neg\i{on}_1.$$
In causal logic, on the other hand, we can express that under the same
assumption {\sl there is a cause} for the light to be off at time 1:
$$\neg\i{on}_1\Ar \i{on}_0\land\i{toggle}.$$
(Performing the toggle action is the cause.) \citeANP{mcc97} showed that
distinctions like this help us solve the frame problem (see Example~5 in
Section~\ref{ssec:cl}) and overcome other difficulties arising in the theory
of reasoning about actions.

The semantics of theories of this kind defines when a propositional
interpretation (truth assignment) is a model of the given theory (is ``causally
explained'' by the theory, in the terminology of  \citeANP{mcc97}).
We do
not reproduce the definition here, because a more general semantics is
described below in Section~\ref{sec:foct}.  But here is an example:
the causal theory
\beq
\ba{rl}
     p&\!\!\!\!\Ar\neg q\\
\neg q&\!\!\!\!\Ar p\\
\ea
\eeq{ex}
has one model, according to the semantics from \cite{mcc97}.  In this
model,~$p$ is true and~$q$ is false.  (Since the bodies of both rules are
true in this model, both rules ``fire''; consequently the heads of
the rules are ``caused''; consequently the truth values of both atoms are
``causally explained.'' This will be discussed formally in
Section~\ref{sec:foct}.)

McCain's translation is applicable to a propositional causal theory~$T$ if
the head of each rule of~$T$ is a literal, and the body is a conjunction of
literals:
\beq
L\Ar A_1\land\cdots\land A_m\land\neg A_{m+1}\land\cdots\neg A_n.
\eeq{r1}
The corresponding logic program consists of the logic programming rules
\beq
L\ar \no\ \neg A_1,\dots,\no\ \neg A_m,\no\ A_{m+1},\dots,\no\ A_n
\eeq{r2}
for all rules~(\ref{r1}) of~$T$.  This program involves two kinds of
negation: negation as failure (\no) and strong, or classical, negation
($\neg$).  According to Proposition~6.7 from
\cite{mcc97c}, complete answer sets of this logic program are identical
to the models of~$T$.  (A set of literals is {\sl complete} if it contains
exactly one member of each complementary pair of literals $A,\neg A$.  We
identify a complete set of literals with the corresponding truth assignment.)

For instance, McCain's translation turns causal theory~(\ref{ex}) into
\beq
\ba{rl}
     p&\ar\no\ q\\
\neg q&\ar \no\ \neg p.\\
\ea
\eeq{ex1}
The only answer set of this program is $\{p,\neg q\}$.  It is complete, and
it corresponds to the model of causal theory~(\ref{ex}).

In this paper we generalize McCain's translation in several ways.  First,
we discard the requirement that the bodies of the given causal rules be
conjunctions of literals.  Second, instead of requiring that the head of
each causal rule be a literal, we allow the heads to be disjunctions of
literals.  In this more general setting, the logic program corresponding to
the given causal theory becomes disjunctive as well.

Third, we study causal rules with heads of the form $L_1\lrar L_2,$
where~$L_1$ and $L_2$ are literals.  Such a rule says that there is a cause
for~$L_1$ and~$L_2$ to be equivalent (``synonymous'') under some condition,
expressed by the body of the rule.  Synonymity rules play an important role
in the theory of commonsense reasoning in view of the fact that humans often
explain the meaning of words by referring to their synonyms.
A synonymity rule
\beq
L_1\lrar L_2\Ar G
\eeq{syn0}
can be translated into logic programming by rewriting it as the pair of rules
$$
\ba l
L_1\lor \o{L_2}\Ar G\\
\o{L_1}\lor L_2\Ar G\\
\ea
$$
($\o L$ stands for the literal complementary to~$L$) and then using our
extension of McCain's translation to rules with disjunctive heads.
It turns out, however, that there is no need to use disjunctive logic
programs in the case of synonymity rules.  If, for instance, $G$
in~(\ref{syn0}) is a literal then the following group of nondisjunctive
rules will do:
$$
\ba l
L_1\ar L_2,\no\ \o{G}\\
L_2\ar L_1,\no\ \o{G}\\
\o{L_1}\ar \o{L_2},\no\ \o{G}\\
\o{L_2}\ar \o{L_1},\no\ \o{G}.
\ea
$$

Finally, we extend the translation from propositional causal rules to
first-order causal rules in the sense of \cite{lif97a}.  This version of
causal logic is useful for defining the semantics of variables in action
descriptions \cite{lif07}.

As part of motivation for our approach to transforming causal theories into
logic programs, we start with a few additional comments on McCain's
translation (Section~\ref{sec:mcc}).  After reviewing the semantics of
causal theories and logic programs in
Sections~\ref{sec:foct} and~\ref{sec:sm}, we describe four kinds of causal
rules that we are interested in and show how to turn a theory consisting of
such rules into a logic program (Section~\ref{sec:main}).
This translation is related to answer set programming in
Section~\ref{sec:gen}, and its soundness is proved in Section~\ref{sec:proof}.

Preliminary reports on this work are published in
\cite{fer06a,fer07,lee10,lif10}.  Some results appear here for the
first time, including the soundness of a representation of a synonymity rule
with variables by a nondisjunctive logic program.

\section{McCain's Translation Revisited} \label{sec:mcc}

\subsection{Incorporating Constraints}

In causal logic, a {\sl constraint} is a rule with the head $\bot$ (falsity).
McCain's translation can be easily extended to constraints with a conjunction
of literals in the body---causal rules of the form
\beq
\bot\Ar A_1\land\cdots\land A_m\land\neg A_{m+1}\land\cdots\land\neg A_n.
\eeq{cons1}
In the language of logic programming,~(\ref{cons1}) can be represented by a
rule similar to~(\ref{r2}):
\beq
\bot\ar \no\ \neg A_1,\dots,\no\ \neg A_m,\no\ A_{m+1},\dots,\no\ A_n.
\eeq{cons2}
Furthermore, each of the combinations $\no\ \neg$ in~(\ref{cons2}) can be
dropped without destroying the validity of the translation; that is to say,
the rule
\beq
\bot\ar A_1,\dots,A_m,\no\ A_{m+1},\dots,\no\ A_n
\eeq{cons3}
can be used instead of~(\ref{cons2}).

\subsection{Eliminating Strong Negation}

As observed in \cite{gel91b},
strong negation can be eliminated from a logic program in favor of additional
atoms.  Denote the new atom representing a negative literal~$\neg A$
by~$\wh A$.  Then~(\ref{r2}) will become
\beq
A_0\ar \no\ \wh{A_1},\dots,\no\ \wh{A_m},\no\ A_{m+1},\dots,\no\ A_n
\eeq{r2a}
if~$L$ is a positive literal~$A_0$, and
\beq
\wh{A_0}
  \ar \no\ \wh{A_1},\dots,\no\ \wh{A_m},\no\ A_{m+1},\dots,\no\ A_n
\eeq{r2b}
if~$L$ is a negative literal~$\neg A_0$.  The {\sl modified McCain
translation} of a causal theory~$T$ consisting of rules of the
forms~(\ref{r1}) and~(\ref{cons1})
includes
\begin{itemize}
\item
rules~(\ref{cons3}) corresponding to the constraints~(\ref{cons1}) of~$T$,
\item
rules~(\ref{r2a}),~(\ref{r2b}) corresponding to the other rules of~$T$, and
\item
the completeness constraints
\beq
\ba l
\ar A,\wh A\\
\ar \no\ A,\no\ \wh A\\
\ea
\eeq{cc}
for all atoms~$A$.
\end{itemize}

For instance, the modified McCain translation of~(\ref{ex}) is
\beq
\ba{rl}
         p&\ar\no\ q\\
\wh q&\ar \no\ \wh p\\
          &\ar p, \wh p\\
          &\ar \no\ p, \no\ \wh p\\
          &\ar q, \wh q\\
          &\ar \no\ q, \no\ \wh q.
\ea
\eeq{ex2}
The only answer set (stable model\footnote{The term ``stable model'' was
introduced in \cite{gel88} to describe the meaning of logic
programs with negation as failure but without strong negation.  When the
stable model semantics was extended to programs with strong negation in
\cite{gel91b}, the term ``answer set'' was proposed as a replacement.})
of this program is $\{p,\wh q\}$.

% This modification is useful to us in view of the fact that the definition
% of a stable model in the context of first-order logic \cite{fer09} is not
% directly applicable to rules with strong negation.

This modification is useful to us in view of the fact that eliminating strong
negation in favor of aditional atoms is part of the definition of a stable
model proposed in \cite[Section~8]{fer09}.

\subsection{Rules as Formulas}

The definition of a stable model for propositional formulas given in
\cite{fer05} and the definition of a stable model for first-order sentences
proposed in \cite{fer09} become generalizations of the original definition
\cite{gel88} when we rewrite rules as logical formulas.
For instance, rules~(\ref{r2a}) and~(\ref{r2b}), rewritten as
propositional formulas, become
\beq
\neg\wh{A_1}\land\cdots\land\neg\wh{A_m}
  \land\neg A_{m+1}\land\cdots\land\neg A_n \rar A_0
\eeq{aux1}
and
\beq
\neg\wh{A_1}\land\cdots\land\neg\wh{A_m}
  \land\neg A_{m+1}\land\cdots\land\neg A_n \rar \wh{A_0}.
\eeq{aux2}
Rule~(\ref{cons3}) can be identified with the formula
\beq
A_1\land\cdots\land A_m\land\neg A_{m+1}\land\dots\land\neg A_n \rar\bot
\eeq{aux3}
or, alternatively, with
\beq
\neg(A_1\land\cdots\land A_m\land\neg A_{m+1}\land\dots\land\neg A_n).
\eeq{aux4}
The completeness constraints for an atom~$A$ turn into the formulas
\beq
\ba l
\neg(A\land\wh A) \\
\neg(\neg A\land\neg \wh A).
\ea
\eeq{cc1}
Here is program~(\ref{ex2}) rewritten in the syntax of propositional logic:
\beq
\ba l
\neg q\rar p\\
\neg\wh p\rar\wh q\\
\neg(p\land\wh p)\\
\neg(\neg p\land\neg \wh p)\\
\neg(q\land\wh q)\\
\neg(\neg q\land\neg \wh q).
\ea
\eeq{ex3}

Note that the process of rewriting a rule as a formula is applicable only
when the rule does not contain strong negation; the symbol~$\neg$
in the resulting formula corresponds to the negation as failure symbol
(\no) in the rule.

One of the advantages of writing rules as formulas is that it allows us to
relate properties of stable models to subsystems of classical logic.  We know,
for instance, that if the equivalence of two sentences can be proved in
intuitionistic logic (or even in the stronger logic of here-and-there)
then these sentences have the same stable models \cite[Theorem~5]{fer09}.
This fact will be used here many times.

% It follows that replacing the completeness
% constraints~(\ref{cc1}) with the intuitionistically equivalent
% formula
% \beq
% \neg(A\lrar\wh A)
% \eeq{cc2}
% does not affect the collection of stable models.
%
% This collection of formulas is essentially identical to logic
% program~(\ref{ex2}), and it has the same stable model.

\subsection{Translating Arbitrary Definite Theories} \label{ssec:tdt}

The requirement, in the definition of McCain's translation, that the bodies
of all causal rules should be conjunctions of literals can be lifted by
slightly modifying the translation process.  Take any set~$T$ of causal
rules of the
forms
\beq
A\Leftarrow G,
\eeq{dr1}
\beq
\neg A\Leftarrow G,
\eeq{dr2}
\beq
\bot\Leftarrow G,
\eeq{dr3}
where $A$ is an atom and~$G$ is an arbitrary propositional formula (rules of
these forms are called {\sl definite\/}).  For each rule~(\ref{dr1}), take the
formula $\neg\neg G\rar A$; for each rule~(\ref{dr2}), the formula
$\neg\neg G\rar \wh A$; for each rule~(\ref{dr3}), the formula~$\neg G$.
Then add completeness constraints~(\ref{cc1}) for all atoms~$A$.
Answer sets of this collection of propositional formulas correspond to the
models of~$T$.

In application to example~(\ref{ex}), this modification of McCain's
translation gives
\beq
\ba l
\neg\neg\neg q\rar p\\
\neg\neg p\rar\wh q\\
\neg(p\land\wh p)\\
\neg(\neg p\land\neg \wh p)\\
\neg(q\land\wh q)\\
\neg(\neg q\land\neg \wh q).
\ea
\eeq{ex4}
It is not surprising that~(\ref{ex4}) has the same answer set as~(\ref{ex3}):
the two collections of formulas are intuitionistically equivalent to each
other.\footnote{Indeed, $\neg\neg\neg q$ is intuitionistically equivalent
to $\neg q$; the equivalence between~$\neg\neg p$ and $\neg\wh p$ is
intuitionistically entailed by the formulas $\neg(p\land\wh p)$ and
$\neg(\neg p\land\neg \wh p)$, which belong both to~(\ref{ex3}) and
to~(\ref{ex4}).}

\section{Review: First-Order Causal Theories} \label{sec:foct}

According to~\cite{lif97a}, a first-order causal theory~$T$ is defined by
\begin{itemize}
\item
a list~$\bp$ of distinct predicate constants,\footnote{We view propositional
symbols as predicate constants of arity~0, so that they are allowed in~\bp.
Equality, on the other hand, may not be declared explainable.\label{ftn}}
called the {\sl explainable symbols} of~$T$,\footnote{To be precise, the
definition in \cite{lif97a} is more general: object and function constants
can be treated as explainable as well.} and
\item
a finite set of {\sl causal rules} of the form $F\Ar G$, where~$F$
and~$G$ are first-order formulas.
\end{itemize}

The semantics of first-order causal theories can be described as
follows. For each $p\in\bp$, choose a new predicate variable~$\v p$ of the
same arity, and let $\v\bp$ stand for the list of all these variables.  By
$T^\dagger(\v\bp)$ we denote the conjunction of the formulas
\beq
\forall\bx(G\rar F^{\scriptsize \bp}_{\scriptsize \v\bp})
\eeq{cr1}
for all rules $F\Ar G$ of~$T$, where~$\bx$ is the list of all free
variables of~$F$,~$G$. (The expression
$F^{\scriptsize \bp}_{\scriptsize \v\bp}$ denotes the result of substituting
the variables $\v\bp$ for the corresponding constants~$\bp$ in~$F$.)

We view~$T$ as shorthand for the sentence
\beq\forall\v\bp(T^\dagger(\v\bp)\lrar(\v\bp=\bp)).\eeq{allct}
(By $\v\bp=\bp$ we denote the conjunction of the formulas
$\forall\bx(\v p(\bx)\lrar p(\bx))$ for all $p\in\bp$, where $\bx$ is a tuple
of distinct object variables.)  Accordingly, by a model of~$T$ we
understand a model of~(\ref{allct}) in the sense of classical logic.
The models of~$T$ are characterized, informally speaking, by the fact that
the interpretation of the explainable symbols~$\bp$ in the model
is the only interpretation of these symbols that is ``causally explained''
by the rules of~$T$.

In the definite case (see Section~\ref{ssec:tdt}) second-order
formula~(\ref{allct}) can be replaced by an equivalent first-order formula
using a process similar to Clark's completion \cite{cla78}, called literal
completion \cite{mcc97}, \cite[Section~5]{lif97a}.  This process is used in
the operation of {\sc CCalc}.

\medskip\noindent{\bf Example 1.}
Let~$T$ be causal theory~(\ref{ex}) with both~$p$
and~$q$ explainable.  Then $T^\dagger(\v p,\v q)$ is
$$(\neg q \rar \v p) \land (p\rar \neg\v q)$$
($\v p$, $\v q$ are propositional variables), so that~$T$ is understood as
shorthand for the second-order propositional formula (``QBF'')
\beq
 \forall (\v p)(\v q)((\neg q \rar \v p) \land (p\rar \neg\v q)
    \lrar (\v p \lrar p) \land (\v q\lrar q)).
\eeq{qbf}
This formula is equivalent to $p\land\neg q$.\footnote{This fact can be
verified by replacing the universal quantifier in~(\ref{qbf}) with the
conjunction of the four propositional formulas obtained by substituting all
possible combinations of values for the variables~$\v p$,~$\v q$, and
simplifying the result.  Alternatively, one can apply literal completion
to rules~(\ref{ex}) and simplify the result.}

\medskip\noindent{\bf Example 2.}
Let~$T$ be the causal theory consisting of two rules:
$$
p(a)\Ar\top
$$
(here $\top$ is the logical constant {\it true}) and
$$
\neg p(x)\Ar\neg p(x),
$$
with the explainable symbol~$p$.
The first rule says that there is a cause for~$a$ to have property~$p$.  The
second rule says that if an object does not have property~$p$ then there is
a cause for that; including this rule in a causal theory has, informally
speaking, the same effect as saying that~$p$ is false by default
\cite[Section~3]{lif97a}.  In this case, $T^\dagger(\v p)$ is
$$\v p(a) \land\forall x(\neg p(x)\rar \neg \v p(x)),$$
so that~$T$ is understood as shorthand for the sentence
$$\forall\v p(\v p(a) \land\forall x(\neg p(x)\rar \neg \v p(x))\lrar
\forall x(\v p(x)\lrar p(x))).$$
This sentence is equivalent to the first-order formula
\beq
\forall x(p(x)\lrar x=a),
\eeq{example2}
as can be verified by applying literal completion to the rules of~$T$.

\section{Review: Stable Models} \label{sec:sm}

Some details of the definition of a stable model proposed in~\cite{fer09}
depend on which propositional connectives are treated as
primitives, and which are viewed as abbreviations.  The convention
there is to take the 0-place connective $\bot$ and the binary
connectives $\land$, $\lor$, $\rar$ as primitives;
$\neg F$ is shorthand for $F\rar\bot$.

In this paper we adopt the view that first-order formulas are formed using a
slightly larger set of propositional connectives:
$$\top,\ \bot,\ \neg,\ \land,\ \lor,\ \rar$$
(as well as the quantifiers $\forall$, $\exists$).  On the other hand, stable
models are only defined here for sentences of a special syntactic form.
A first-order sentence is a {\sl rule}\footnote{Or {\sl program rule},
to distinguish it from causal rules in the sense of Section~\ref{sec:foct}.}
if it has the form $\uc(F\rar G)$ and has no occurrences of~$\rar$ other than
the one explicitly shown.\footnote{$\uc F$ stands for the universal closure
of~$F$.}  If a sentence~$F$ does not contain implication then we will identify
it with the rule $\top\rar F$.  For instance, propositional
formulas~(\ref{aux1})--(\ref{ex3}) are rules.
% In a rule of the form
% $\uc F$,~$F$ is the {\sl head} of the rule; in a rule of the form
% $\uc(F\rar G)$,~$F$ is the {\sl body} and~$G$ is the {\sl head}.
A {\sl logic program} is a conjunction of rules.  The definition of a
stable model below is more limited than the
definition from \cite{fer09} because it is only applicable to programs,
not to arbitrary sentences.  For instance, it does
not cover the formulas $(p\rar q)\rar r$ and $(p\rar q)\lor r$.  On the other
hand, it is simpler than the general definition, and it is sufficient
for our present purposes.

We need the following notation from~\cite{lif93e}.
If $p$ and $q$ are predicate constants of the same arity then $p\leq q$ stands
for the formula
$$\forall {\bf x}(p({\bf x})\rar q({\bf x})),$$
where {\bf x} is a tuple of distinct object variables.
If {\bf p} and {\bf q} are tuples $p_1,\dots,p_n$ and $q_1,\dots,q_n$ of
predicate constants then ${\bf p}\leq{\bf q}$ stands for the conjunction
$$(p_1\leq q_1) \land \cdots \land (p_n\leq q_n),$$
and ${\bf p}<{\bf q}$ stands for
$({\bf p}\leq{\bf q}) \land\neg({\bf q}\leq{\bf p})$.
In second-order logic, we apply the same notation to tuples of
predicate variables.

Let~${\bf p}$ be a list of distinct predicate constants; members of~{\bf p}
will be called {\sl intensional predicates}.\footnote{This list usually
consists of all predicate symbols occurring in the heads of rules; those are
the predicates that we ``intend to characterize'' by the rules of the
program.  The original definition of a stable model \cite{gel88} treats
all predicates as intensional.}
For each $p\in\bp$, choose a predicate variable~$\v p$ of the
same arity, and let $\v\bp$ stand for the list of all these variables.
For any logic program $F$, by $\sm_{\bf p}[F]$ we
denote the second-order sentence
\beq
F \land \neg \exists \v\bp ((\v\bp<\bp) \land F^\diamond(\v\bp)),
\eeq{smd}
where $F^\diamond(\v\bp)$ is the formula obtained from~$F$ by replacing, for
every $p\in\bp$, each occurrence of~$p$ that is not in the scope of negation
with $\v p$.
A model of $F$ is {\sl stable} (relative to the set~${\bf p}$ of intensional
predicates) if it satisfies $\sm_{\bf p}[F]$.\footnote{We can make two
comments about the relation of this treatment of stable models to earlier
work.  First, if we drop from the definition of $\sm$ the words ``that is not
in the scope of negation'' then it will turn into the definition of parallel
circumscription \cite{mcc86,lif85}.  It follows that if a logic program
does not contain negation then the class of its stable models is
identical to the class of its minimal models.  The stipulation in the
description of $F^\diamond(\v\bp)$ that intensional predicates
in the scope of negation are not replaced by variables is a reflection of the
idea of negation as failure.  Second, the
operator of $\sm$ as defined in \cite{fer09} produces, in application to
a logic program, a second-order formula that is usually more complex
than~(\ref{smd}) but is  equivalent to it.}

\medskip\noindent{\bf Example 3.}  Let~$F$ be the propositional formula
$\neg p\rar q$
(the one-rule program $q\ar\no\ p$, in traditional notation).
If both~$p$ and~$q$ are intensional then $F^\diamond(\v p,\v q)$ is
$$\neg p\rar \v q,$$
so that $\sm_{pq}[F]$ is
$$(\neg p\rar q)\land\neg\exists (\v p)(\v q) (((\v p,\v q)<(p,q))\land
 (\neg p\rar \v q)).$$
This formula is equivalent to $\neg p\land q$.\footnote{Methods for
simplifying the result of applying the operator $\sm$ are discussed in
~\cite{fer09}.}  Consequently~$F$ has one stable model: $p$ is false and~$q$
is true.

\medskip\noindent{\bf Example 4.}
Let~$F$ be the formula
\beq
\forall x(\neg p(x)\rar (q(x)\lor\neg q(x)))
\eeq{example4}
(it can be thought of as a formula representation of the {\sc lparse} choice
rule\break\verb+{q(X)} :- not p(X)+).\footnote{This rule would not be accepted by
{\sc lparse}, however, because it is ``nonrestricted.'' For a description of
the language of {\sc lparse} see
{\tt http://www.tcs.hut.fi/Software/ smodels/lparse.ps}.\label{ftlp}}
If we take~$q$ to be the only intensional predicate then~$F^\diamond(\v q)$ is
$$
\forall x(\neg p(x)\rar(\v q(x)\lor\neg q(x))).
$$
Consequently $\sm_q[F]$ is
$$
\forall x(\neg p(x)\rar (q(x)\lor\neg q(x)))
\land\,\neg\exists \v q((\v q<q)
\land\forall x(\neg p(x)\rar(\v q(x)\lor\neg q(x)))).
$$
The first conjunctive term here is logically valid and can be dropped.  The
second is equivalent to the first-order formula
$\neg\exists x(p(x)\land q(x))$,
which reflects the intuitive meaning of the choice rule above:~$q$ is an
arbitrary set disjoint from~$p$.

The relationship between the definition of a stable model given above and
the operation of answer set solvers is discussed in Section~\ref{sec:gen}.

If programs~$F$ and~$G$ are intuitionistically equivalent then
$\sm_{\bf p}[F]$ is equivalent to $\sm_{\bf p}[G]$, that is to say,~$F$ and~$G$
have the same stable models.  Moreover, for establishing that~$F$ and~$G$
have the same stable models we only need to derive~$F\lrar G$
intuitionistically from the excluded middle formulas $\uc(H\lor\neg H)$ for
some formulas~$H$ that do not contain intensional predicates.  This fact
follows from \cite[Theorem~5]{fer09}.

\section{Turning a Causal Theory into a Logic Program}\label{sec:main}

\subsection{Four Types of Causal Rules} \label{ssec:types}

In the rest of the paper, we assume that the bodies of causal rules do not
contain implication.  This is not an essential limitation, because in
classical logic $\rar$ can be expressed in terms of other connectives,
and the meaning of a causal rule does not change if we replace its body (or
head) by a classically equivalent formula.

Here are four types of rules that we are going to consider,
in the order of increasing complexity of their heads:
\begin{itemize}
\item
The head is $\bot$, that is, the rule is a constraint.  Such causal rules will
be also called {\sl C-rules}.
\item
The head is a literal containing an explainable predicate symbol.
These are {\sl L-rules}.
\item
The head has the form $L_1\lrar L_2$, where each~$L_i$ is a literal
containing an explainable predicate symbol.  These are {\sl synonymity
rules}, or {\sl S-rules}.
\item The head has the form $L_1\lor\cdots\lor L_n$ ($n\geq 0$), where
each~$L_i$ is a literal containing an explainable predicate symbol.
These are {\sl D-rules}.
\end{itemize}

All C-rules and L-rules can be viewed also as D-rules, and any S-rule
can be replaced with an equivalent pair of D-rules (see Lemma~\ref{lem:s} in
Section~\ref{ssec:proof2}).  Nevertheless, we give special attention here to
rules of the first three types,  and the reason is that our translation
handles such rules in special ways.  It appears that causal rules of
types~C,~L, and~S will be more important than general D-rules in applications
of this work to the automation of reasoning about actions.

On the other hand, the possibility of reducing types~C,~L, and~S to type~D
plays an important role in the proof of the soundness of our translation
(Section~\ref{sec:proof}).  This is one of the reasons why we are interested
in general D-rules.

The requirement, in the definitions of types L, S and D, that the literals
in the head of the rule contain explainable predicate symbols is not an
essential limitation.  If, for instance, the predicate symbol in the head of
$L\Ar G$ is not explainable then this rule can be equivalently replaced by the
C-rule $\bot\Ar G\land\o{L}$.  If a rule has the form
$$L_1\lrar L_2\Ar G$$
and the predicate symbol in~$L_1$ is not explainable then the rule can be
replaced by
$$
\ba l
L_2\Ar G\land L_1,\\
\o{L_2}\Ar G\land \o{L_1}.
\ea
$$
If a rule has the form
$$L_1\lor\cdots\lor L_n\Ar G$$
and the predicate symbol in~$L_1$ is not explainable then the rule can be
replaced by
$$L_2\lor\cdots\lor L_n\Ar G\land\o{L_1}.$$

\subsection{Translating C-Rules and L-Rules} \label{ssec:cl}

The transformation described in this section generalizes McCain's translation,
in the form described in Section~\ref{ssec:tdt}, to first-order causal
theories.

The operator $\tr_c$, which transforms any C-rule into a program rule,
is defined by the formula
$$\tr_c[\bot\Ar G]=\uc\neg G.$$

The operator $\tr_l$, which transforms any L-rule into a program rule,
is defined by the formulas
$$
\ba {rl}
\tr_l[p({\bf t})\Ar G]=&\!\!\!\uc(\neg\neg G\rar p({\bf t})),\\
\tr_l[\neg p({\bf t})\Ar G]=&\!\!\!\uc(\neg\neg G\rar \wh{p}({\bf t}))
\ea
$$
({\bf t} is a tuple of terms).

If~$T$ is a causal theory consisting of C-rules and L-rules then its
translation $\tr[T]$ is the logic program obtained by conjoining
\begin{itemize}
\item
the rules obtained by applying $\tr_c$ to the C-rules of~$T$,
\item
the rules obtained by applying $\tr_l$ to the L-rules of~$T$, and
\item
the completeness constraints
\beq
\ba l
\forall\bx\neg(p(\bx)\land\wh p(\bx)),\\
\forall\bx\neg(\neg p(\bx)\land\neg \wh p(\bx))
\ea
\eeq{cc3}
($\bx$ is a tuple of distinct object variables) for all explainable
predicate symbols~$p$ of~$T$.
\end{itemize}

Let~$\bp$ be the list of explainable predicate symbols~$p$ of~$T$, and
let~$\wh \bp$ be the list of the corresponding predicate symbols~$\wh p$.
Take the union of~$\bp$ and~$\wh \bp$ to be the set of intensional predicates.
 Then the stable models of the logic program $\tr[T]$ are ``almost identical''
to the models of~$T$; the difference is due to the fact that the language
of~$T$ does not contain the symbols~$\wh\bp$.  Let~\i{CC} be
the conjunction of all completeness constraints~(\ref{cc3}).  Then the
relationship between~$T$ and~$\tr[T]$ can be described as follows:
\beq
\sm_{\scriptsize \bp\bph}[\tr[T]]\;\hbox{\sl is equivalent to }T\land\i{CC}.
\eeq{soundness}
This claim, expressing the soundness of our translation, is extended in
Sections~\ref{ssec:s} and~\ref{ssec:d} to causal
theories containing S-rules and D-rules, and its proof is given in
Section~\ref{sec:proof}.

Since the conjunction of formulas~(\ref{cc3}) is classically equivalent to
\beq
\forall\bx(\wh p(\bx)\lrar\neg p(\bx)),
\eeq{c4}
sentence \i{CC} can be viewed as the conjunction of explicit definitions of the
predicates~$\wh \bp$ in terms of the predicates~$\bp$.  Consequently the
relationship~(\ref{soundness}) shows that $\sm_{\scriptsize \bp\bph}[\tr[T]]$
is a definitional extension of~$T$.  The models of $\tr[T]$ that are
stable relative to $\bp\wh\bp$ can be characterized as the models of~$T$
extended by the interpretations of the predicates $\wh\bp$ that are provided by
definitions~(\ref{c4}).

\medskip\noindent{\bf Example 1, continued.}  If~$T$ is causal
theory~(\ref{ex}) with both~$p$ and~$q$ explainable then $\tr[T]$ is
the conjunction of formulas~(\ref{ex4}).  The result of applying the operator
$\sm_{pq\wh p\wh q}$ to this conjunction is equivalent to
$$p\land\neg q\land\neg \wh p\land\wh q.$$
Recall that~$T$ is equivalent to the first half of this conjunction
(Section~\ref{sec:foct}).  The second half tells us that the truth values
of~$\wh p$,~$\wh q$ are opposite to the truth values of~$p$,~$q$.
In the only stable model of~(\ref{ex4}),~$p$ and~$\wh q$ are true, and
$\wh p$ and~$q$ are false; if we ``forget'' the truth values
of~$\wh p$ and~$\wh q$ then we will arrive at the model
of~(\ref{ex}).

\medskip\noindent{\bf Example 2, continued.}  Our translation
turns the causal theory from Example~2 into the conjunction of the rules
$$
\ba l
\neg\neg\top\rar p(a),\\
\forall x(\neg\neg\neg p(x)\rar{\wh p}(x)),\\
\forall x \neg(p(x) \land {\wh p}(x)),\\
\forall x \neg(\neg p(x) \land \neg {\wh p}(x)),\\
\ea
$$
or, after intuitionistically equivalent transformations,
$$
\ba l
p(a),\\
\forall x(\neg p(x)\rar{\wh p}(x)),\\
\forall x \neg(p(x) \land {\wh p}(x)),\\
\forall x \neg(\neg p(x) \land \neg {\wh p}(x)).
\ea
$$
The result of applying $\sm_{p\wh p}$ to the conjunction of these formulas
is equivalent to the conjunction of~(\ref{example2}) with the formula
$\forall x(\wh p(x)\lrar \neg p(x))$, which says that~$\wh p$ is the complement
of~$p$.

\medskip\noindent{\bf Example 5.}
Consider the following causal rules:
\beq
\ba{rl}
\i{on}_1(x)      &\!\!\!\Ar \i{toggle}(x) \land \neg\i{on}_0(x),\\
\neg \i{on}_1(x) &\!\!\!\Ar \i{toggle}(x) \land \i{on}_0(x),\\
\i{on}_1(x)      &\!\!\!\Ar \i{on}_0(x) \land \i{on}_1(x),\\
\neg\i{on}_1(x)  &\!\!\!\Ar \neg\i{on}_0(x) \land \neg\i{on}_1(x).
\ea
\eeq{toggle}
The first pair of rules describes the effect of toggling a switch~$x$:
this action causes the fluent $\i{on}(x)$ at time~1 to take the value opposite
to its value at time~0.  The second pair solves the frame problem
\cite{sha97} for the
fluent $\i{on}(x)$ by postulating that if the value of that fluent at time~1
is equal to its previous value then there is a cause for this.  (Inertia, in
the sense of commonsense reasoning, is the cause.)
Let~$T$ be the causal theory with rules~(\ref{toggle}) and with ~$\i{on}_1$
as the only explainable symbol.  Using literal
completion, we can check that~$T$ is equivalent to
\beq
\forall x(\i{on}_1(x) \lrar
((\i{on}_0(x) \land \neg\i{toggle}(x))\lor
(\neg\i{on}_0(x) \land \i{toggle}(x)))).
\eeq{toggle-res}
Our translation turns~$T$ into the conjunction of the rules
\beq
\ba l
\forall x(\neg\neg(\i{toggle}(x) \land \neg\i{on}_0(x)) \rar \i{on}_1(x)),\\
\forall x(\neg\neg(\i{toggle}(x) \land \i{on}_0(x)) \rar \wh{\i{on}_1}(x)),\\
\forall x(\neg\neg(\i{on}_0(x) \land \i{on}_1(x))\rar \i{on}_1(x)),\\
\forall x(\neg\neg(\neg\i{on}_0(x)\land\neg\i{on}_1(x))\rar\wh{\i{on}_1}(x)),\\
\forall x\neg(\i{on}_1(x)\land\wh{\i{on}_1}(x)),\\
\forall x\neg(\neg\i{on}_1(x)\land\neg\wh{\i{on}_1}(x)),\\
\ea
\eeq{toggle-lp-prep}
or, equivalently,\footnote{Removing the
double negations in the first two lines of~(\ref{toggle-lp-prep}) is
possible because neither \i{toggle} nor $\i{on}_0$ is intensional
(see the comment on equivalent transformations of logic programs at the end of
Section~\ref{sec:sm}).  In a similar way, the antecedent of the third
impication in~(\ref{toggle-lp-prep}) can be replaced by
$\i{on}_0(x) \land \neg\neg\i{on}_1(x)$; the equivalence between
$\neg\neg\i{on}_1(x)$ and $\neg \wh{\i{on}_1}(x)$ is intuitionistically
entailed by the last two lines of~(\ref{toggle-lp-prep}).  The fourth line
of~(\ref{toggle-lp-prep}) is simplified in a similar way.}
\beq
\ba l
\forall x(\i{toggle}(x) \land \neg\i{on}_0(x) \rar \i{on}_1(x)),\\
\forall x(\i{toggle}(x) \land \i{on}_0(x) \rar \wh{\i{on}_1}(x)),\\
\forall x(\i{on}_0(x) \land \neg \wh{\i{on}_1}(x)\rar \i{on}_1(x)),\\
\forall x(\neg\i{on}_0(x)\land\neg\i{on}_1(x)\rar\wh{\i{on}_1}(x)),\\
\forall x\neg(\i{on}_1(x)\land\wh{\i{on}_1}(x)),\\
\forall x\neg(\neg\i{on}_1(x)\land\neg\wh{\i{on}_1}(x)).
\ea
\eeq{toggle-lp}
The result of applying $\sm_{\scriptsize{\i{on}_1\wh{\i{on}_1}}}$ to this
program is equivalent to the conjunction of~(\ref{toggle-res}) with the
formula $\forall x(\wh{\i{on}_1}(x)\lrar\neg\i{on}_1(x))$, which says
that~$\wh{\i{on}_1}$ is the complement of~$\i{on}_1$.

\medskip\noindent{\bf Example 6.}  The constraint
$$\bot\Ar \i{toggle}(\i{badswitch})$$
expresses that \i{badswitch} is stuck: the action of toggling it is not
executable.  If we add this constraint to the causal theory from
Example~5 then the rule
$$\neg\i{toggle}(\i{badswitch})$$
will be added to its translation~(\ref{toggle-lp}).

\medskip
The bodies of causal rules in Examples~5 and~6 are syntactically simple:
they are conjunctions of literals.  The general definitions of a C-rule and
an L-rule do not impose any restrictions on the form of the body, and in
applications of causal logic to formalizing commonsense knowledge this
generality is often essential.  For instance, the statement ``each position
must have at least one neighbor'' in the landscape structure of the Zoo
World\footnote{The challenge of formalizing the Zoo World was proposed as
part of the Logic Modelling Workshop
({\tt http:/www/ida.liu.se/ext/etai/lmw/}).  The possibility of addressing
this challenge using {\sc CCalc} is discussed in \cite[Section~4]{akm04}.}
would be represented in causal logic by a C-rule with a quantifier in the
body.

\subsection{Translating S-Rules} \label{ssec:s}

We will turn now to translating synonymity rules (Section~\ref{ssec:types}).
The operator~$\tr_s$, transforming any such rule into a logic
program, is defined by the formulas
$$
\ba l
\tr_s[p_1({\bf t}^1)\lrar p_2({\bf t}^2)\Ar G]
=\tr_s[\neg p_1({\bf t}^1)\lrar\neg p_2({\bf t}^2)\Ar G]\\
\qquad =\uc(\neg\neg G \land p_1({\bf t}^1)\rar p_2({\bf t}^2))
       \land\uc(\neg\neg G \land p_2({\bf t}^2)\rar p_1({\bf t}^1))\,\land\\
\qquad\quad\uc
(\neg\neg G \land \wh{p_1}({\bf t}^1)\rar \wh{p_2}({\bf t}^2))
\land\uc
(\neg\neg G \land \wh{p_2}({\bf t}^2)\rar \wh{p_1}({\bf t}^1)),\\
\ea
$$
$$
\ba l
\tr_s[\neg p_1({\bf t}^1)\lrar p_2({\bf t}^2)\Ar G]
=\tr_s[p_1({\bf t}^1)\lrar\neg p_2({\bf t}^2)\Ar G]\\
\qquad =\uc(\neg\neg G \land \wh{p_1}({\bf t}^1)\rar p_2({\bf t}^2))
       \land\uc(\neg\neg G \land p_2({\bf t}^2)\rar \wh{p_1}({\bf t}^1))\,\land\\
\qquad\quad\uc
(\neg\neg G \land p_1({\bf t}^1)\rar \wh{p}_2({\bf t}^2))
\land\uc
(\neg\neg G \land \wh{p}_2({\bf t}^2)\rar p_1({\bf t}^1))
\ea
$$
(${\bf t}^1$, ${\bf t}^2$ are tuples of terms).  The definition of program
$\tr[T]$ from Section~\ref{ssec:cl} is extended to causal theories that may
contain S-rules, besides C-rules and L-rules, by adding that $\tr[T]$ includes
also
\begin{itemize}
\item
the rules obtained by applying $\tr_s$ to the S-rules of~$T$.
\end{itemize}

\medskip\noindent{\bf Example 7.}  Extend the theory from Example~5 by the
rule
\beq
\i{dark} \lrar \neg \i{on}_1(myswitch)\Ar\top,
\eeq{dark}
where \i{dark} is explainable.  The corresponding logic program is obtained
from~(\ref{toggle-lp}) by adding the rules
\beq
\ba l
\wh{\i{dark}} \rar \i{on}_1(myswitch),\\
\i{on}_1(myswitch)\rar\wh{\i{dark}},\\
\i{dark} \rar \wh{\i{on}_1}(myswitch),\\
\wh{\i{on}_1}(myswitch)\rar \i{dark},\\
\neg(\i{dark}\land\wh{\i{dark}}),\\
\neg(\neg\i{dark}\land\neg\wh{\i{dark}}).
\ea
\eeq{dark-lp}

We will see that the soundness property~(\ref{soundness}) holds for
arbitary causal theories consisting of rules of types~C,~L, and~S.

\subsection{Translating D-Rules} \label{ssec:d}

A D-rule (Section~\ref{ssec:types}) has the form
\beq
\bigvee_{\scriptsize A\in\pa}A\lor\bigvee_{\scriptsize A\in\na}\neg A
\Ar G
\eeq{cr}
for some sets $\pa$,~$\na$ of atomic formulas.

If~$A$ is an atomic
formula~$p({\bf t})$, where $p\in\bp$ and {\bf t} is a tuple of terms,
then by~$\wh A$ we will denote the formula $\wh{p}({\bf t})$.
The operator~$\tr_d$ transforms D-rule~(\ref{cr}) into the program rule
\beq
\widetilde\forall\left(\,\neg\neg
G\land\bigwedge_{\scriptsize
A\in\pa}(\wh{A}\lor\neg\wh{A})\land\bigwedge_{\scriptsize
A\in\na}(A\lor\neg A)\;\rar\,\bigvee_{\scriptsize A\in\pa}
A\lor\bigvee_{\scriptsize A\in\na} \wh{A}\,\right).\eeq{lr}

\medskip\noindent{\bf Example 8.}  The result of applying~$\tr_d$ to the D-rule
$$p\lor\neg q\lor\neg r\Ar s$$
is
$$
\neg\neg s\land (\wh p\lor\neg\wh p)\land (q\lor\neg q)\land(r\lor\neg r)
   \rar p\lor \wh q \lor \wh r.
$$
\medskip

The number of ``excluded middle formulas'' conjoined with $\neg\neg G$
in~(\ref{lr}) equals the number of disjunctive terms in the head of
D-rule~(\ref{cr}).  In particular, if~(\ref{cr}) is an L-rule then the
antecedent of~(\ref{lr}) contains one such formula.  For instance, in
application to the first rule of~(\ref{ex}) $\tr_d$ produces the program rule
$$\neg\neg\neg q\land (\wh p\lor\neg\wh p) \rar p,$$
which is more complex than the first rule of~(\ref{ex4}).

For a fixed collection~$\bp$ of explainable symbols, let~$C$,~$L$,~$S$, and~$D$
be finite sets of causal rules of types~C,~L,~S, and~D respectively.  By
$\tr[C,L,S,D]$ we denote the logic program obtained by conjoining
\begin{itemize}
\item
the rules obtained by applying $\tr_c$ to all rules from~$C$,
\item
the rules obtained by applying $\tr_l$ to all rules from~$L$,
\item
the programs obtained by applying $\tr_s$ to all rules from~$S$,
\item
the rules obtained by applying $\tr_d$ to all rules from~$D$,
\item
the completeness constraints (\ref{cc3}) for all explainable symbols~$p$.
\end{itemize}
Our most general form of the soundness theorem, proved in
Section~\ref{sec:proof}, asserts that
\beq
\sm_{\scriptsize \bp\bph}[\tr[C,L,S,D]]\;\hbox{\sl is equivalent to }
T\land\i{CC}
\eeq{soundness-g}
for the causal theory~$T$ with the set of rules $C\cup L\cup S\cup D$.  In
the special case when~$D$ is empty this theorem turns into the assertion
stated at the end of Section~\ref{ssec:s}.

\section{Using Answer Set Solvers to Generate Models of a Causal Theory}
\label{sec:gen}

The discussion of answer set solvers in this section, as almost any
discussion of software, is somewhat informal.  We assume here that the
first-order language under consideration does not contain function constants
of nonzero arity.

An answer set solver can be viewed as a system for generating
stable models in the sense of Section~\ref{sec:sm}, with three caveats.  First,
currently available solvers require that the input program have a syntactic
form that is much more restrictive than the syntax of first-order
logic.\footnote{They also require that the input satisfy some safety
conditions.  See, for instance, Chapter~3 of the {\sc dlv} manual,
{\tt http://www.dbai.tuwien.ac.at/proj/dlv/man/.}\label{ftn:safe}}
Preprocessing based on intuitionistically equivalent
transformations often helps us alleviate this difficulty. There exists a
tool, called {\sc f2lp} \cite{lee09a}, that converts first-order formulas of a
rather general kind into logic programs accepted by {\sc lparse}.
The rules produced by the process described in the previous section
have no existential quantifiers in their heads, and all quantifiers in
their bodies are in the scope of negation. Consequently, these rules satisfy
a syntactic condition that guarantees the correctness of the translation
implemented in {\sc f2lp}.

Second, answer set solvers represent stable models by sets of ground atoms.
To introduce such a representation, we usually choose a finite set of object
constants that includes all object constants occurring in the program, and
restrict attention to Herbrand interpretations of the extended language.  The
\verb+#domain+ construct of {\sc lparse}\footnote{See
Footnote~$(^{\ref{ftlp}})$.} can be used to specify the object constants
constituting the domain of the variables in the program.

Third, most existing answer set solvers are unaware of the possibility of
non-intensional (or {\sl extensional\/}) predicates.  Treating a predicate
constant as extensional can be simulated using a choice rule
\cite[Theorem~2]{fer09}.  There is also another approach to overcoming
this limitation.  Take a conjunction~$E$ of some
ground atoms containing extensional predicates, and assume that we are
interested in the Herbrand stable models of a program~$F$ that interpret the
extensional predicates in accordance with~$E$ (every atom from~$E$ is true;
all other atoms containing extensional predicates are false).  Under some
syntactic conditions,\footnote{Specifically, under the assumption that every
occurrence of every extensional predicate in~$F$ is in the scope of negation
or in the antecedent of an implication.} these stable models are identical to
the Herbrand stable models
of~$F\land E$ with all predicate constants treated as intensional.  This
can be proved using the splitting theorem from \cite{fer09a}.

\medskip\noindent{\bf Example 4, continued.}
We would like to find the stable models of~(\ref{example4}), with~$q$
intensional, that have the universe $\{a,b,c,d\}$ and
make~$p$ true on~$a$, $b$ and false on~$c$, $d$.  This is the same as
to look for the Herbrand stable models of the formula
$$\forall x(\neg p(x)\rar (q(x)\lor\neg q(x)))\land p(a)\land p(b),$$
with~$c$ and~$d$ viewed as object constants of the language along with~$a$
and~$b$, and with both~$p$ and~$q$ taken to be intensional.

A representation of this example in the language of {\sc lparse} is
shown in Figure~\ref{fig1}.
\begin{figure}[t]
\hrule\bigskip

\flushleft{\tt
u(a;b;c;d).

\#domain u(X).

\{q(X)\} :- not p(X).

p(a;b).
}

\bigskip\hrule
\caption{Example~4 with a 4-element universe in the language of {\sc lparse}}
\label{fig1}
\end{figure}
The auxiliary predicate {\tt u} describes the universe of the
interpretations that we are interested in.  The first line is
shorthand for
$$\hbox{\tt u(a). u(b). u(c). u(d).}$$
and the last line is understood by {\tt lparse} in a similar way.

Given this input, the answer set solver {\sc smodels} generates 4 stable
models, representing the subsets of $\{a,b,c,d\}$ that are disjoint from
$\{a,b\}$:
\begin{verbatim}
Answer: 1
Stable Model: p(b) p(a) u(d) u(c) u(b) u(a)
Answer: 2
Stable Model: p(b) p(a) q(d) u(d) u(c) u(b) u(a)
Answer: 3
Stable Model: p(b) p(a) q(c) u(d) u(c) u(b) u(a)
Answer: 4
Stable Model: p(b) p(a) q(d) q(c) u(d) u(c) u(b) u(a)
\end{verbatim}

\medskip
In application to the logic program obtained from a causal theory~$T$ as
described in Section~\ref{sec:main}, this process often allows us
to find the models of~$T$ with a given universe and given extents of
extensional predicates.

\medskip\noindent{\bf Example 7, continued.}  There are two switches,
\i{myswitch} and \i{hisswitch}.  It is dark in my room at time~1 if
and only if \i{myswitch} is not \i{on} at time~1.  At time~0, both switches
are \i{on}; then \i{hisswitch} is toggled, and \i{myswitch}
is not.  Is it dark in my room at time~1?  We would like to answer this
question using answer set programming.

This example of commonsense reasoning involves inertia (the value of the
fluent $\i{on}(\i{myswitch})$ does not change because this fluent
is not affected by the action that is executed) and indirect effects of
actions: whether or not it is dark in the room at time~1 after performing
some actions is determined by the effect of these actions on the fluent
$\i{on}(\i{myswitch})$.

Mathematically, we are talking here about the causal theory~$T$ with
rules~(\ref{toggle}) and~(\ref{dark}), with the object
constant~\i{hisswitch} added to the language, and with the explainable
symbols $\i{on}_1$ and $\i{dark}$.  We are interested in the Herbrand models
of~$T$ in which the extents of the extensional
predicates are described by the atoms
$$
\i{on}_0(\i{myswitch}),\ \i{on}_0(\i{hisswitch}),\ \i{toggle}(\i{hisswitch}).
$$

 As we have seen, the logic program $\tr[T]$ is equivalent to the conjunction
of rules~(\ref{toggle-lp}) and~(\ref{dark-lp}).  The corresponding
{\sc lparse} input file is shown in Figure~\ref{fig2}.  In this file, the
``true negation'' symbol {\tt -} is used in the ASCII representations of the
symbols $\wh{\i{on}_1}$ and $\wh{\i{dark}}$; the {\sc lparse} counterparts of
the rules
$$
\ba l
\forall x\neg(\i{on}_1(x)\land\wh{\i{on}_1}(x)),\\
\neg(\i{dark}\land\wh{\i{dark}})\\
\ea
$$
are dropped, because such ``coherence'' conditions are verified by
the system automatically.

\begin{figure}[t]
\hrule\bigskip

\flushleft{\tt
u(myswitch;hisswitch).

\#domain u(X).
\bigskip

on1(X) :- toggle(X), not on0(X).

-on1(X) :- toggle(X), on0(X).

on1(X) :- on0(X), not -on1(X).

-on1(X) :- not on0(X), not on1(X).

:- not on1(X), not -on1(X).
\bigskip

on1(myswitch) :- -dark.

-dark :- on1(myswitch).

-on1(myswitch) :- dark.

dark :- -on1(myswitch).

:- not dark, not -dark.
\bigskip

on0(myswitch;hisswitch).

toggle(hisswitch).
}

\bigskip\hrule
\caption{Example~7 with two switches in the language of {\sc lparse}}
\label{fig2}
\end{figure}

Given this input, {\sc smodels} generates the only model of~$T$ satisfying
the given conditions:
\begin{verbatim}
Answer: 1
Stable Model: -on1(hisswitch) on1(myswitch) -dark toggle(hisswitch)
on0(hisswitch) on0(myswitch) u(hisswitch) u(myswitch)
\end{verbatim}
The presence of {\tt -dark} in this model
tells us that it is not dark in the room at time~1.
\medskip

The example above is an example of ``one-step temporal
projection''---predicting the value of a fluent after performing a single
action in a given state.  Some other kinds of temporal reasoning and planning
can be performed by generating models of simple modifications of the given
causal theory \cite[Section~3.3]{giu04}; this is one of the ideas behind the
design of {\sc CCalc} and {\sc coala}.  McCain's translation reviewed in the
introduction and its generalization presented in Section~\ref{sec:main} allow
 us to solve such problems automatically using an answer set solver.

\section{Proof of Soundness} \label{sec:proof}

To prove claim~(\ref{soundness-g}), which expresses the soundness of our
translation, we will first establish it for the case when $C=L=S=\emptyset$
(Section~\ref{ssec:proof1}).  In this ``leading special case'' all rules of
the given causal theory are D-rules, and they are converted to program rules
using the translation $\tr_d$.  Then we will derive the soundness theorem in
full generality (Section~\ref{ssec:proof2}).

\subsection{Leading Special Case} \label{ssec:proof1}

Let~$T$ be a finite set of causal rules of the form~(\ref{cr}).  Let~$\Pi$
be the conjunction of the corresponding program rules~(\ref{lr}),
and let \i{CC}, as before, stand for the conjunction of the completeness
constraints (\ref{cc3}) for all explainable symbols~$p$ of~$T$.  We want to
show that
\beq
\sm_{\scriptsize \bp\bph}[\Pi\land\i{CC}]\;\hbox{\sl is equivalent to }
T\land\i{CC}.
\eeq{leading}
% Since \i{CC} has no strictly positive occurrences of intensional predicates,
% $\sm_{\scriptsize\bp\bph}[\Pi\land\i{CC}]$ is equivalent to
% $
% \sm_{\scriptsize\bp\bph}[\Pi]\land\i{CC}
% $
% \cite[Section~5.1]{fer09}.  Therefore the leading special case of the
% soundness theorem that we want to establish is equivalent to the claim that
% \i{CC} entails
% % \beq
% $$\sm_{\scriptsize\bp\bph}[\Pi]\lrar T.$$
% % \eeq{claim}

The key steps in the proof below are
Lemma~\ref{lem:5} (one half of the equivalence) and
Lemma~\ref{lem:8} (the other half).

In the statement of the following lemma, $\neg\bp$ stands for the list of
predicate expressions\footnote{See \cite[Section~3.1]{lif93e}.}
$\lambda\bx\neg p(\bx)$, where $\bx$ is a list of distinct object variables,
for all~$p$ from~$\bp$.  By $\v\bp$, $\v\bph$ we denote the lists of predicate
variables used in the second-order formula
$\sm_{\scriptsize\bp\bph}[\Pi\land\i{CC}]$ (see Section~\ref{sec:sm}).

\begin{lemma}\label{lem:1}
Formula $(\v\bp,\v\bph)<(\bp,\neg\bp)$ is equivalent to
$$\bigvee_{\scriptsize
p\in\bp}\left(\left((\v\bp,\v\wh{\bp})\le(\bp,\neg\bp)\right) \land
\exists\bx(\neg \v p(\bx)\land\neg\v\wh{p}(\bx)\right)).$$
\end{lemma}

%\medskip\noindent{\bf Proof.}
\begin{proof}
 Note first that
$$
% \ba{rl}
\ba l
(\v\bp,\v\bph)<(\bp,\neg\bp)\\
\;\llrar
\left((\v\bp,\v\wh{\bp})\le(\bp,\neg\bp)\right)
  \land\neg\left((\bp,\neg\bp)\le(\v\bp,\v\wh{\bp})\right)\\
\;\llrar
\left((\v\bp,\v\wh{\bp})\le(\bp,\neg\bp)\right)
  \land\bigvee_{\scriptsize p\in\bp}
   \exists\bx((p(\bx)\land\neg\v p(\bx))
     \lor(\neg p(\bx)\land\neg\v\wh{p}(\bx)))\\
\;\llrar
\bigvee_{\scriptsize p\in\bp}(\left((\v\bp,\v\wh{\bp})\le(\bp,\neg\bp)\right)
  \land
   \exists\bx((p(\bx)\land\neg\v p(\bx))
     \lor(\neg p(\bx)\land\neg\v\wh{p}(\bx)))).\\
\ea
$$
The disjunction after $\exists\bx$ is equivalent to
\beq
     (p(\bx)\lor \neg\v\wh{p}(\bx))
  \land(\neg\v p(\bx)\lor\neg p(\bx))
  \land(\neg\v p(\bx)\lor\neg\v\wh{p}(\bx)).
\eeq{conj}
Since $(\v\bp,\v\wh{\bp})\le(\bp,\neg\bp)$ entails
$$\v p(\bx)\rar p(\bx)\hbox{ and }\v \wh{p}(\bx)\rar \neg p(\bx),$$
the first conjunctive term of~(\ref{conj}) can be rewritten as
$\neg\v\wh{p}(\bx)$, and the second term as $\neg\v p(\bx)$, so
that~(\ref{conj}) will turn into $\neg\v p(\bx)\land\neg\v\wh{p}(\bx)$.
\end{proof}

% \subsection{Proof of Sufficiency}
For any formula~$F$, by $F_{\Sigma 1}$ we denote the formula
$$F^{\scriptsize (\v\bp)(\v\bph)}
   _{\scriptsize (\v\bp\land \bp)(\neg\v\bp\land\neg\bp)}$$
where $\v\bp\land \bp$ is understood as the list of predicate expressions
$$\lambda\bx(\v p(\bx)\land p(\bx))$$
for all $p\in\bp$, and $\neg\v\bp\land\neg\bp$ is understood in a similar
way.\footnote{For the definition of $F^{\scriptsize \bp}_{\scriptsize \v\bp}$
see Section~\ref{sec:foct}.}

\begin{lemma}\label{lem:2}
Formula
$$\left((\v\bp,\v\bph)<(\bp,\neg\bp)\right)_{\Sigma 1}$$
is equivalent to $\v\bp\neq\bp$.
\end{lemma}

\begin{proof} In view of Lemma \ref{lem:1},
$\left((\v\bp,\v\bph)<(\bp,\neg\bp)\right)_{\Sigma 1}$ is equivalent to
the disjunction of the formulas
 \beq
\ba{rcl}
\left(\bigwedge_{\scriptsize p\in\bp}\forall\bx(\v p(\bx)\rar
p(\bx))_{\Sigma 1}\right)
&\land& \left(\bigwedge_{\scriptsize
p\in\bp}\forall\bx(\v\wh{p}(\bx)\rar
\neg p(\bx))_{\Sigma 1}\right)\\
&\land&  \exists\bx(\neg \v p(\bx)\land
\neg\v\wh{p}(\bx))_{\Sigma 1}
\ea
\eeq{lemma2.1}
for all $p\in\bp$. It is easy to verify that
$$
\ba{rl} (\v p(\bx)\rar p(\bx))_{\Sigma 1}&=\;\;
(\v p(\bx)\land p(\bx)\rar p(\bx))\;\;\llrar\;\;\top\;,\\ \\
(\v\wh{p}(\bx)\rar \neg p(\bx))_{\Sigma 1}&=\;\;(\neg \v
p(\bx)\land \neg p(\bx)\rar \neg
p(\bx))\;\;\llrar\;\;\top,\\ \\
(\neg \v p(\bx)\land\neg\v\wh{p}(\bx))_{\Sigma 1}
&\llrar\;\;((\neg \v p(\bx)\lor\neg p(\bx))\land\neg(\neg \v
p(\bx)\land\neg p(\bx)))\\
&\llrar\;\; (\v p(\bx)\lrar\neg p(\bx))\\
&\llrar\;\; \neg(\v p(\bx)\lrar p(\bx)).\\
\ea
$$
Therefore (\ref{lemma2.1}) is equivalent to
$\exists\bx \neg (\v p(\bx)\lrar p(\bx))$,
so that the disjunction of all formulas~(\ref{lemma2.1})
is equivalent to $\v\bp\neq \bp$.
\end{proof}

 If~$A$ is an atomic formula~$p({\bf t})$, where
$p\in\bp$ and {\bf t} is a tuple of terms, then we will write
$\v A$ for $\v p({\bf t})$,
and $\wh{A}$ for $\v \wh{p}({\bf t})$.  By $\ouc F$ we denote the
formula $\forall\bx F$, where~$\bx$ is list of all free object variables of~$F$
(``object-level universal closure'').

Define $H(\v\bp,\v\bph)$ to be the conjunction of the implications
\beq
\ouc\!\left(
G\rar\bigvee_{\scriptsize A\in\pa}((\v\wh{A}\lor A)\rar \v
A)\lor\bigvee_{\scriptsize A\in \na}((\v A\lor\neg A)\rar \v\wh{A})\right)
\eeq{lr1}
for all rules (\ref{cr}) in $T$.

\begin{lemma}\label{lem:3}
Formula $\sm_{\scriptsize \bp\bph}[\Pi\land\i{CC}]$ is equivalent to
\beq
\Pi\land\i{CC}\land \forall (\v\bp)(\v\bph)(((\v\bp,\v\bph)<(\bp,\neg\bp))\rar
\neg H(\v\bp,\v\bph)).
\eeq{suf}
\end{lemma}

\begin{proof}
Every occurrence of every intensional predicate in \i{CC} is in the scope
of a negation.  Consequently
$\sm_{\scriptsize \bp\bph}[\Pi\land \i{CC}]$ is
$$\Pi\land\i{CC}\land
\neg\exists(\v\bp)(\v\bph)(((\v\bp,\v\bph)<(\bp,\bph))\land
\Pi^\diamond(\v\bp,\v\bph)\land\i{CC}),
$$
which is equivalent to
$$\Pi\land\i{CC}\land\forall(\v\bp)(\v\bph)(((\v\bp,\v\bph)<(\bp,\neg\bp))
   \rar\neg\Pi^\diamond(\v\bp,\v\bph)).
$$
We will conclude the proof by showing that \i{CC} entails
$$\Pi^\diamond(\v\bp,\v\bph)\lrar H(\v\bp,\v\bph).$$
The left-hand side of this equivalence is the conjunction of the formulas
$$%\beq
\ouc\!\left(
\neg\neg G\land\bigwedge_{\scriptsize A\in\pa}(\v\wh{A}\lor \neg\wh{A})
\land\bigwedge_{\scriptsize A\in\na}(\v A\lor\neg A)
\rar\bigvee_{\scriptsize A\in\pa} \v A\lor\bigvee_{\scriptsize A\in\na}\v\wh{A}
\right)
$$%\eeq{lemma4.3}
for all rules (\ref{cr}) in $T$.  Under the assumption \i{CC} this formula
can be rewritten as
$$\ouc\!\left( G\rar
\bigvee_{\scriptsize A\in\pa}\neg(\v\wh{A}\lor
A)\lor\bigvee_{\scriptsize A\in\na}\neg(\v A\lor\neg
A)\lor\bigvee_{\scriptsize A\in \pa}\v A\lor\bigvee_{\scriptsize
A\in\na} \v\wh{A}\right).$$
The last formula is equivalent to
$$
\ouc\!\left(G\rar \bigvee_{\scriptsize A\in\pa}(\neg(\v\wh{A}\lor
A)\lor \v A)\lor\bigvee_{\scriptsize A\in\na}(\neg(\v A\lor\neg
A)\lor\v\wh{A})\right).
$$
and consequently to~(\ref{lr1}).
\end{proof}

\begin{lemma}\label{lem:4}
$T^\dagger(\v\bp)$ is equivalent to $H(\v\bp,\v\bph)_{\Sigma 1}$.
\end{lemma}

\begin{proof}
Formula $T^\dagger(\v\bp)$ is the conjunction of the formulas
\beq\ouc\!\left( G\rar \bigvee_{\scriptsize A\in\pa} \v
A\lor\bigvee_{\scriptsize A\in\na}\neg \v A\right) \eeq{dag}
for all rules~(\ref{cr}) in~$T$. On the other hand,
$H(\v\bp,\v\bph)_{\Sigma 1}$ is the conjunction of the formulas
\beq\ouc\!\left( G\rar\bigvee_{\scriptsize A\in\pa}((\v\wh{A}\lor
A)\rar \v A)_{\Sigma 1}\lor\bigvee_{\scriptsize A\in\na}((\v
A\lor\neg A)\rar\v\wh{A})_{\Sigma 1}\right) \eeq{lr1s} for all
rules (\ref{cr}) in $T$. It remains to observe that
$$
\ba{rl} ((\v\wh{A}\lor A)\rar \v A)_{\Sigma 1}
  & =\;\;  (\neg\v A\land \neg A)\lor A\rar \v A\land A \\
  & \llrar\;\; \neg \v A\lor A\rar \v A \land A\\
  & \llrar\;\; (\v A\land \neg A)\lor (\v A\land A)\\
  & \llrar\;\; \v A,
\ea
$$
and that, similarly, $((\v A\lor\neg A)\rar\v\wh{A})_{\Sigma 1}$ is
equivalent to $\neg \v A$.
\end{proof}

\begin{lemma}\label{lem:5}
% For a causal theory $T$ and a set of explainable symbols $\bp$, the
% following holds: \beq
$\sm_{\scriptsize\bp\bph}[\Pi\land\i{CC}]\models T\land\i{CC}$.
%\eeq{th2.1}
\end{lemma}

\begin{proof}
Recall that, according to Lemma~\ref{lem:3},
$\sm_{\scriptsize\bp\bph}[\Pi\land\i{CC}]$ is equivalent to~(\ref{suf}).
The second conjunctive term of~(\ref{suf}) is \i{CC}.
The first conjunctive term is equivalent to~$T^\dagger(\bp)$. From the other
two terms we conclude:
$$
\forall\v\bp(((\v\bp,\v\bph)<(\bp,\bph))_{\Sigma 1}\rar\neg H(\v\bp,\v\bph)_{\Sigma 1}).
$$
By Lemma \ref{lem:2} and Lemma \ref{lem:4}, this formula is equivalent to
$$
\forall\v\bp((\v\bp\neq\bp)\rar\neg T^\dagger(\v\bp)),
$$
and consequently to
$$
\forall\v\bp(T^\dagger(\v\bp)\rar (\v\bp=\bp)).
$$
The conjunction of the last formula
with~$T^\dagger(\bp)$ is equivalent to~(\ref{allct}).
\end{proof}

% \subsection{Proof of Necessity}

For any formula~$F$, by $F_{\Sigma 2}$ we denote the formula
$$F^{\scriptsize \v\bp}
   _{\scriptsize (((\v\bp,\v\bph)\le(\bp,\neg\bp))
                 \land\neg\v\bp\land\neg\v\bph)\lrar\neg\bp}$$
where the subscript
$$(((\v\bp,\v\bph)\le(\bp,\neg\bp))
                 \land\neg\v\bp\land\neg\v\bph)\lrar\neg\bp$$
is understood as the list of predicate expressions
$$\lambda\bx((((\v\bp,\v\bph)\le(\bp,\neg\bp))
             \land\neg\v p(\bx)\land\neg\v\wh{p}(\bx))\lrar\neg p(\bx))$$
for all $p\in\bp$.

\begin{lemma}\label{lem:6}
Formula
$$(\v\bp\neq\bp)_{\Sigma 2}$$
is equivalent to $(\v\bp,\v\bph)<(\bp,\neg\bp)$.
\end{lemma}

\begin{proof}
Formula
$(\v\bp\neq\bp)_{\Sigma 2}$ is equivalent to
$$
\bigvee_{\scriptsize p\in\bp}
  \exists\bx(\v p(\bx)\lrar\neg p(\bx))_{\Sigma 2}
$$
that is,
$$
\bigvee_{\scriptsize p\in\bp}\exists\bx((((\v\bp,\v\bph)\le
(\bp,\neg\bp))\land\neg\v p(\bx)\land\neg \v\wh{p}(\bx)\lrar \neg
p(\bx))\lrar \neg p(\bx)).
$$
This formula can be equivalently rewritten as
$$
\bigvee_{\scriptsize p\in\bp}(((\v\bp,\v\bph)\le
(\bp,\neg\bp))\land\exists\bx(\neg\v p(\bx)\land\neg
\v\wh{p}(\bx))),
$$
which is equivalent to $(\v\bp,\v\bph)<(\bp,\neg\bp)$ by
Lemma \ref{lem:1}.
\end{proof}

\begin{lemma}\label{lem:7}
The implication
$$
(\v\bp,\v\bph)\leq(\bp,\neg\bp)\rar
(T^\dagger(\v\bp)_{\Sigma2}\lrar  H(\v\bp,\v\bph))
$$
is logically valid.
\end{lemma}

\begin{proof*}
Recall that $T^\dagger(\v\bp)$ is the conjunction of
implications~(\ref{dag}) for all rules~(\ref{cr}) in~$T$.
Consequently $T^\dagger(\v\bp)_{\Sigma 2}$ is the conjunction of the
formulas
$$
\ouc\!\left(G\rar \bigvee_{\scriptsize A\in\pa} (\v A)_{\Sigma 2} \lor
\bigvee _{\scriptsize A\in\na}\neg(\v A)_{\Sigma 2}\right),
$$
that is to say,
$$
 \ba{rl}
\ouc( G\rar
     &\bigvee_{\scriptsize A\in\pa} ((((\v\bp,\v\bph)\le
      (\bp,\neg\bp))\land\neg \v A\land\neg\v\wh{A})\lrar\neg A)\,\lor\\
     &\bigvee_{\scriptsize A\in\na} \neg ((((\v\bp,\v\bph)\le
      (\bp,\neg\bp))\land\neg \v A\land\neg\v\wh{A})\lrar\neg A).
%          &\bigvee_{\scriptsize p\in\na} ((((\bp,\bph)\le
%(\bp,\neg\bp))\land\neg \v A\land\neg\v\wh{A})\lrar A)).
\ea
$$
Under the assumption
\beq
(\v\bp,\v\bph)\le(\bp,\neg\bp)
\eeq{le}
the last formula can be equivalently rewritten as
$$
\ouc\!\left(G\rar \bigvee_{\scriptsize A\in\pa} ((\v
A\lor\v\wh{A})\lrar A)\lor\bigvee_{\scriptsize A\in\na} ((
\v A\lor\v\wh{A})\lrar \neg A)\right).$$ It remains to check that,
under assumption~(\ref{le}), \beq (\v A\lor\v\wh{A})\lrar A
\eeq{f1} can be equivalently rewritten as \beq \v\wh{A}\lor
A\rar \v A, \eeq{f2} and \beq \v A\lor\v\wh{A}\lrar \neg A
\eeq{f3} can be rewritten as \beq \v A\lor \neg A\rar\v\wh{A}.
\eeq{f4} Formula~(\ref{f1}) is equivalent to \beq (\v A\rar A)\land
(\v \wh{A}\rar A)\land (A\rar \v A\lor \v \wh{A}).
\eeq{f4a} Since assumption~(\ref{le}) entails $\v A\rar A$ and $\v
\wh{A}\rar \neg A$, formula~(\ref{f4a}) can be rewritten as
\beq \neg \v\wh{A}\land (A\rar \v A). \eeq{f5}
On the other hand, formula~(\ref{f2}) is equivalent to
$$(\v\wh{A}\rar \v A)\land (A\rar \v A),$$
which, under assumption~(\ref{le}), can be rewritten as~(\ref{f5}) as well.
In a similar way, each of the formulas~(\ref{f3}),~(\ref{f4}) can be
transformed into
$$\neg \v A\land (\neg A\rar \v \wh{A}).\mathproofbox$$
\end{proof*}

\begin{lemma}\label{lem:8}
$T\land\i{CC}\models \sm_{\scriptsize\bp\bph}[\Pi\land\i{CC}]$.
\end{lemma}

\begin{proof}
Recall that~$T$ is equivalent to
\beq
T^\dagger(\bp)\land\forall\v\bp(T^\dagger(\v\bp)\rar(\v\bp=\bp)).
\eeq{pr8a}
Since the first conjunctive term is equivalent to~$\Pi$, $T\land\i{CC}$
entails
\beq
\Pi\land\i{CC}.
\eeq{pr8b}
From the second conjunctive term of~(\ref{pr8a}) we conclude
$$
T^\dagger(\v\bp)_{\scriptsize\Sigma 2}\rar(\v\bp=\bp)_{\scriptsize\Sigma 2}
$$
and consequently
$$
\forall(\v\bp)(\v\bph)((\v\bp\neq\bp)_{\scriptsize\Sigma 2}\rar \neg
T^\dagger(\v\bp)_{\scriptsize\Sigma 2}).
$$
By Lemma \ref{lem:6}, this is equivalent to
$$
\forall(\v\bp)(\v\bph)(((\v\bp,\v\bph)<(\bP,\neg\bP))\rar \neg
T^\dagger(\v\bp)_{\scriptsize\Sigma 2})$$
and, by Lemma \ref{lem:7}, to
$$
\forall(\v\bp)(\v\bph)(((\v\bp,\v\bph)<(\bp,\neg\bp))\rar \neg
 H(\v\bp,\v\bph)).
$$
By Lemma~\ref{lem:3}, the conjunction of this formula with~(\ref{pr8b})
is equivalent to sentence $\sm_{\scriptsize\bp\bph}[\Pi\land\i{CC}]$.
\end{proof}

Assertion~(\ref{leading}) follows from Lemmas~\ref{lem:5} and~\ref{lem:8}.

\subsection{General Case} \label{ssec:proof2}

\begin{lemma}\label{lem:c}
For any C-rule~$R$, $\tr_c[R]$ is intuitionistically equivalent to $\tr_d[R]$.
\end{lemma}

\begin{proof}
If~$R$ is $\bot\Ar G$ then $\tr_c[R]$ is $\uc\neg G$, and $\tr_d[R]$ is
$\uc(\neg\neg G\rar\bot)$.
\end{proof}

\begin{lemma}\label{lem:l}
For any L-rule~$R$, the conjunction~\i{CC} of completeness constraints
intuitionistically entails
$$\tr_l[R]\lrar\tr_d[R].$$
\end{lemma}

\begin{proof}
If~$R$ is $p({\bf t})\Ar G$ then $\tr_l[R]$ is
$$
\uc(\neg\neg G\rar p({\bf t})),
$$
and $\tr_d[R]$ is
$$
\uc(\neg\neg G\land(\wh p({\bf t}) \lor \neg\wh p({\bf t}))
   \rar p({\bf t})).
$$
Since \i{CC} intuitionistically entails
\beq
\neg(p({\bf t})\lrar \wh p({\bf t})),
\eeq{cc5}
it is sufficient to check that~$p({\bf t})$ can be derived from~(\ref{cc5}) and
\beq
\wh p({\bf t}) \lor \neg\wh p({\bf t})\rar p({\bf t})
\eeq{fe1}
by the deductive means of intuitionistic propositional logic.
Since~(\ref{fe1}) is equivalent to $p(\bf t)$ in classical propositional
logic, it is easy to see that~$\neg\wh p({\bf t})$ can be derived
from~(\ref{cc5}) and~(\ref{fe1}) in classical propositional logic.  By
Glivenko's theorem,\footnote{This theorem \cite{gli29},
\cite[Theorem~3.1]{min00} asserts that if a
formula beginning with negation can be derived from a set~$\Gamma$ of formulas
in classical propositional logic then it can be derived from~$\Gamma$ in
intuitionistic propositional logic as well.} it follows that it can be
derived intuitionistically as well.  Since~$p({\bf t})$ is intuitionistically
derivable from~(\ref{fe1}) and~$\neg\wh p({\bf t})$,
we can conclude that~$p({\bf t})$ is intuitionistically derivable
from~(\ref{cc5}) and~(\ref{fe1}).

The case when~$R$ is $\neg p({\bf t})\Ar G$ is similar.
\end{proof}

\begin{lemma}\label{lem:s}
If~$R$ is an S-rule
\beq
L_1\lrar L_2\Ar G
\eeq{s1}
and~$R_1$, $R_2$ are the D-rules
\beq
L_1\lor \o{L_2}\Ar G\ \hbox{ and }\ \o{L_1}\lor L_2\Ar G
\eeq{s2}
then the conjunction~\i{CC} of completeness constraints
intuitionistically entails
$$\tr_s[R]\lrar \tr_d[R_1]\land \tr_d[R_2].$$
\end{lemma}

\begin{proof}
If each of the literals~$L_i$ is an atom~$A_i$ then $\tr_s[R]$ is the
conjunction of the formulas
\beq
\ba l
\uc(\neg\neg G\land A_1\rar A_2),\\
\uc(\neg\neg G\land A_2\rar A_1),\\
\uc(\neg\neg G\land \wh{A_1}\rar \wh{A_2}),\\
\uc(\neg\neg G\land \wh{A_2}\rar \wh{A_1}),
\ea
\eeq{tr-r}
$\tr_d[R_1]$ is
\beq
\uc(\neg\neg G \land (\wh{A_1}\lor\neg\wh{A_1})\land (A_2\lor\neg A_2)
           \rar A_1\lor \wh{A_2}),
\eeq{tr-r1}
and $\tr_d[R_2]$ is
\beq
\uc(\neg\neg G \land (A_1\lor\neg A_1)\land (\wh{A_2}\lor\neg \wh{A_2})
           \rar \wh{A_1}\lor A_2).
\eeq{tr-r2}
We need to show that \i{CC} intuitionistically entails the equivalence
between the conjunction of formulas~(\ref{tr-r}) and the conjunction of
formulas~(\ref{tr-r1}),~(\ref{tr-r2}).  Since~\i{CC} intuitionistically
entails
\beq \neg({A_1}\lrar\wh{{A_1}}) \eeq{a1}
and
\beq \neg({A_2}\lrar\wh{{A_2}}), \eeq{a2}
it is sufficient to check that the conjunction of
formulas~(\ref{a1}),~(\ref{a2}),
\beq {A_1}\lrar {A_2}\eeq{c1}
and
\beq \wh{{A_1}}\lrar\wh{{A_2}}\eeq{c2}
is equivalent in intuitionistic propositional logic to the conjunction of
formulas~(\ref{a1}),~(\ref{a2}),
\beq (\wh{{A_1}}\lor\neg\wh{{A_1}})\land({A_2}\lor\neg {A_2})\rar
{A_1}\lor\wh{{A_2}}\eeq{b1}
and
\beq ({A_1}\lor\neg {A_1})\land (\wh{{A_2}}\lor\neg\wh{{A_2}})\rar
\wh{{A_1}}\lor {A_2}.
\eeq{b2}

{\it Left-to-right:}  Assume (\ref{a1})--(\ref{c2}) and
\beq
(\wh{{A_1}}\lor\neg\wh{{A_1}})\land({A_2}\lor\neg{A_2});
\eeq{twodis}
our goal is to derive intuitionistically ${A_1}\lor\wh{A_2}$.  Consider two
cases, in accordance with the first disjunction in~(\ref{twodis}).
{\it Case 1: $\wh{A_1}$.} Then, by (\ref{c2}), $\wh{{A_2}}$, and
consequently ${A_1}\lor\wh{{A_2}}$.  {\it Case 2: $\neg\wh{A_1}$.}
Consider two cases, in accordance with the second disjunction
in~(\ref{twodis}).  {\it Case 2.1: $A_2$}.  Then, by (\ref{c1}), $A_1$, and
consequently ${A_1}\lor\wh{{A_2}}$. {\it Case 2.2: $\neg A_2$.}  Then,
by (\ref{c1}), $\neg {A_1}$, which contradicts~(\ref{a1}).

Thus we proved that (\ref{b1}) is intuitionistically derivable from
(\ref{a1})--(\ref{c2}). The proof for (\ref{b2}) is similar.

{\it Right-to-left:} Let~$\Gamma$ be the set consisting of
formulas~(\ref{a1}),~(\ref{a2}),~(\ref{b1}),~(\ref{b2}) and $A_1$.  We claim
that~$A_2$ can be derived from~$\Gamma$ in intuitionistic propositional logic.
Note that, classically,
\begin{itemize}
\item Formula (\ref{a1}) is equivalent to ${A_1}\lrar\neg\wh{A_1}$,
\item Formula (\ref{a2}) is equivalent to ${A_2}\lrar\neg\wh{A_2}$, and
\item Formula (\ref{b2}) is equivalent to $\wh{A_1}\lor {A_2}$.
\end{itemize}
It follows that $\neg \wh{A_2}$ is derivable from~$\Gamma$ in classical
propositional logic. By Glivenko's theorem, it follows that~$\neg\wh{A_2}$ is
derivable from~$\Gamma$ intuitionistically as well.  Hence the antecedent
of~(\ref{b2}) is an intuitionistic consequence of~$\Gamma$, and so is the
consequent $\wh{A_1}\lor {A_2}$.  In combination with $A_1$ and~(\ref{a1}),
this gives us $A_2$.

We conclude that~$A_1\rar A_2$ is intuitionsistically derivable
from~(\ref{a1}), (\ref{a2}), (\ref{b1}) and (\ref{b2}).  The derivability of
the implication $A_2\rar A_1$ from these formulas can be proved in a similar
way.  Thus~(\ref{c1}) is an intuitionistic consequence of~(\ref{a1}),
(\ref{a2}), (\ref{b1}), and~(\ref{b2}).

The derivability of~(\ref{c2}) from these formulas in propositional
intuitionistic logic is proved in a similar way.

The cases when the literals~$L_i$ are negative, or when one of them is
positive and the other is negative, are similar.
\end{proof}

\noindent{\it Proof of the soundness property~(\ref{soundness-g})}.
Let~$C$,~$L$,~$S$, and~$D$ be sets of causal rules of types~C,~L,~S, and~D
respectively, and let~$T$ be the causal theory with the set of rules
$C\cup L\cup S\cup D$.  Consider the causal theory~$T'$ obtained from~$T$
by replacing each rule~(\ref{s1}) from~$S$ with the corresponding
rules~(\ref{s2}). According to the result~(\ref{leading}) of
Section~\ref{ssec:proof1},
$$
\sm_{\scriptsize \bp\bph}[\Pi\land\i{CC}]\;\hbox{ is equivalent to }
T'\land\i{CC},
$$
where~$\Pi$ is the conjunction of the program rules~$\tr_d[R]$ for all
rules~$R$ of~$T'$.  It is clear that $\Pi\land\i{CC}$ is $\tr[T']$, and
that~$T'$ is equivalent to~$T$.  Consequently
\beq
\sm_{\scriptsize \bp\bph}[\tr[T']]\;\hbox{ is equivalent to } T\land\i{CC}.
\eeq{fin1}
On the other hand, Lemmas~\ref{lem:c},~\ref{lem:l} and~\ref{lem:s} show that
the formulas $\tr[T']$ and $\tr[C,L,S,D]$ are intuitionistically
equivalent to each other, because each of them contains~\i{CC} as a conjunctive
term.  It follows that
\beq
\sm_{\scriptsize \bp\bph}[\tr[T']]\hbox{ is equivalent to }
\sm_{\scriptsize \bp\bph}[\tr[C,L,S,D]].
\eeq{fin2}
Assertion~(\ref{soundness-g}) follows from~(\ref{fin1}) and~(\ref{fin2}).
 % \qed

\section{Conclusion}

In this paper we generalized McCain's embedding of definite causal theories
into logic programming.  We expect that this work will provide a theoretical
basis for extending the system {\sc coala} to more expressive action languages,
including the modular action language MAD \cite{ren09}.  It is essential,
from this perspective, that our translation is applicable to synonymity rules,
because such rules are closely related to the main new feature of MAD, its
{\bf import} construct.

Our translation is not applicable to causal rules with quantifiers in the
head.  It may be possible to extend it to positive occurrences of existential
quantifiers, since an existentially quantified formula can be thought of as
an infinite disjunction.  But the translation would be a
formula with positive occurrences of existential quantifiers as well, and
it is not clear how to turn such a formula into executable code.

In the future, we would like to extend the translation described above to
causal theories with explainable function symbols, which
correspond to non-Boolean fluents in action languages.  Since the definition
of a stable model does not allow function symbols to be intensional, such a
generalization would have to involve extending the language by auxiliary
predicate symbols.

\section*{Acknowledgements}

We are grateful to the anonymous referees for useful comments.
Joohyung Lee was partially supported by the National Science Foundation
under grant IIS-0916116 and by the Office of the
Director of National Intelligence (ODNI), Intelligence Advanced Research
Projects Activity (IARPA), through US Army.  Yuliya Lierler was supported by
a 2010 Computing Innovation Fellowship.  Vladimir Lifschitz and Fangkai Yang
were supported by the National Science Foundation under grant IIS-0712113.
All statements of fact,
opinion or conclusions contained herein are those of the authors and
should not be construed as representing the official views or policies
of IARPA, the ODNI or the U.S.~Government.

\bibliographystyle{acmtrans}
\bibliography{bib}
\end{document}